\documentclass[10pt,twocolumn,letterpaper]{article}

\usepackage{cvpr}
\usepackage{times}
\usepackage{epsfig}
\usepackage{graphicx}
\usepackage{amsmath}
\usepackage{amssymb}
\usepackage{subcaption}
\usepackage{authblk}

\usepackage[breaklinks=true,bookmarks=false]{hyperref}
\cvprfinalcopy 

\begin{document}

\title{Deep Face Recognition Model Compression via Knowledge Transfer and Distillation}
\author{Jayashree Karlekar$^1$, Jiashi Feng$^2$, Zi Sian Wong$^1$, Sugiri Pranata}
\affil[1]{Panasonic R\&D Center Singapore}
\affil[2]{National University of Singapore}
\affil[ ]{\textit {\{karlekar.jayashree,zisian.wong,sugiri.pranata\}@sg.panasonic.com}}
\affil[ ]{\textit {elefjia@nus.edu.sg}}

\maketitle

\begin{abstract}
Fully convolutional networks (FCNs) have become de facto tool to achieve very high-level performance for many vision and non-vision tasks in general and face recognition in particular. Such high-level accuracies are normally obtained by very deep networks or their ensemble. However, deploying such high performing models to resource constraint devices or real-time applications is challenging. In this paper, we present a novel model compression approach based on student-teacher paradigm for face recognition applications. The proposed approach consists of training teacher FCN at bigger image resolution while student FCNs are trained at lower image resolutions than that of teacher FCN.  We explored three different approaches to train student FCNs:  knowledge transfer (KT), knowledge distillation (KD) and their combination. Experimental evaluation on LFW and IJB-C datasets demonstrate comparable improvements in accuracies with these approaches. Training low-resolution student FCNs from higher resolution teacher offer fourfold advantage of accelerated training, accelerated inference, reduced memory requirements and improved accuracies. We evaluated all models on IJB-C dataset and achieved state-of-the-art results on this benchmark. The teacher network and some student networks even achieved Top-1 performance on IJB-C dataset. The proposed approach is simple and hardware friendly, thus enables the deployment of high performing face recognition deep models to resource constraint devices.
\end{abstract}

\section{Introduction}
\label{section_intro}
Recently, deep networks \cite{bookbengio} have achieved state-of-the-art accuracies on many tasks ranging from computer vision to natural language processing. Several competitions, such as image classification, object detection and semantic segmentation \cite{imagenet,voc,coco} are annually held to push algorithmic advancement to achieve top accuracies without any restriction on computational and memory resources. Most of the time, the top performing approaches in these competitions use very deep networks or their ensemble, comprising millions of parameters making them memory and compute intensive.
Deployment of such winning model(s) without compromising performance is challenging.

To address issues of deep network deployability, 
many different approaches are proposed in literature to compress over parametrized and compute intensive deep models.
Many competitions are also launched recently in which restrictions are imposed on model size and/or inference time 
\cite{frvt11ongoing,lowpower,youtube8m}.
Primarily, there exists two main strategies to compress deep networks:
one is to design new fast architectures such as MobileNets \cite{mobilenet} or another is to compress existing high accuracy deep models. In this paper, we focus on latter approach.   
Yu et. al \cite{compressionsurvey} presents a comprehensive survey of techniques to compress high accuracy deep models. Most popular among these techniques are parameter/channel pruning, parameter quantization, low-rank factorization and knowledge distillation. Parameter/channel pruning, low-rank factorization and quantization techniques focus on reducing model size and computational requirements while keeping network architecture same, whereas, distillation approaches modify network architecture to reduce memory and computational requirements. In this paper, we present our approach based on knowledge transfer and distillation framework without pruning, quantization, factorization or modifying architecture to reduce memory and computational resources for face recognition applications. 

\begin{figure}[t!]
\centering
\includegraphics[height=2.3in]{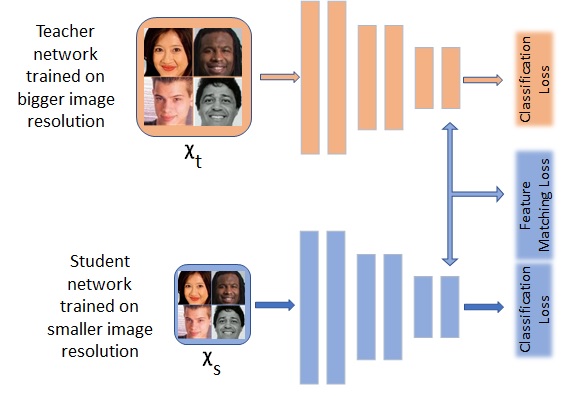}
\caption{In proposed teacher-student approach, teacher and student networks share same architecture while using different input image resolutions.}
\label{fig_kd}
\vspace{-4mm}
\end{figure}

Among many computer vision tasks, face recognition has achieved very high accuracies on various datasets such as LFW \cite{lfwdata}, IJB-A \cite{ijba}, IJB-C \cite{ijbc} and MegaFace \cite{megaface}. However, high  accuracies are normally achieved with very deep fully convolutional networks (FCNs) \cite{arcface,l2loss,crystalloss,faceiconicity} or their ensemble \cite{covariate} trained with huge datasets publicly available such as CASIA \cite{casia}, MS-Celeb1M \cite{msceleb} and VGGFace2 \cite{vggface2}. To compress these FCNs, we present a novel approach based on student-teacher paradigm for face recognition applications. The proposed approach consists of training teacher FCN at higher image resolution while student FCNs are trained at lower image resolutions than that of teacher FCN as illustrated in Figure \ref{fig_kd}. In this setting, both teacher and student networks share same architecture and have exactly same number of parameters. Model acceleration is achieved implicitly with reduced input image resolution rather than reduced model parameters (Section \ref{section_motivation}).

In proposed approach, as student and teacher networks share same deep architecture, we explored three different techniques to train student network: 1) via knowledge transfer (KT) \cite{net2net}, 2) via knowledge distillation (KD) \cite{distillation} and 3) via their combination, so that it outperforms itself had it been trained from scratch without assistant from teacher network. 
Both KD and KT techniques are based on teacher-student paradigm, however they extract ''knowledge'' differently from teacher to train student network. 
In KD \cite{distillation} technique, knowledge is distilled from various layers of powerful teacher network in the form of ``layer representations'' to train less powerful student network. While in KT \cite{net2net} paradigm, ''knowledge'' of teacher in the form of ``layer parameters'' is used to train student network.
By combining both these paradigms to train low-resolution student FCNs we achieve fourfold advantage of accelerated training, accelerated inference, reduced memory requirements and improved accuracies (Sections \ref{section_approach}). We evaluated all our models on IJB-C dataset and achieved state-of-the-art results on this benchmark. Moreover, our teacher network and some student networks achieved Top-1 accuracies on IJB-C dataset.
The proposed approach is simple and hardware friendly, thus enabling the deployment of high performing face recognition deep models to resource constraint devices (Section \ref{section_results}).   

\section{Deep Face Model Compression with KD and KT}
\subsection{Motivation}
\label{section_motivation}

\begin{figure}[t!]
    \centering
    \begin{subfigure}[t]{0.21\textwidth}
        \centering
        \includegraphics[width=\textwidth]{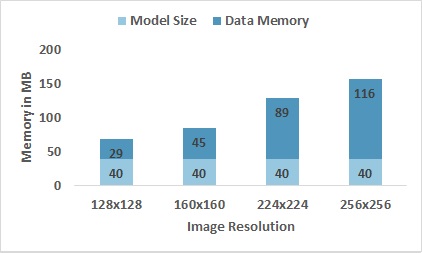}
        \caption{Memory}
    \end{subfigure} 
    \hspace{5mm}
    \begin{subfigure}[t]{0.21\textwidth}
        \centering
        \includegraphics[width=\textwidth]{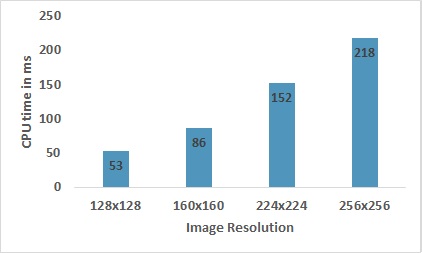}
        \caption{Time}
    \end{subfigure}
    \caption{Total (a) memory and (b) time requirements of Inception-BN model for different input resolution images.}
\label{fig_gglmodel_time}
\end{figure}

\begin{figure}[t!]
\centering
\includegraphics[height=1.2in]{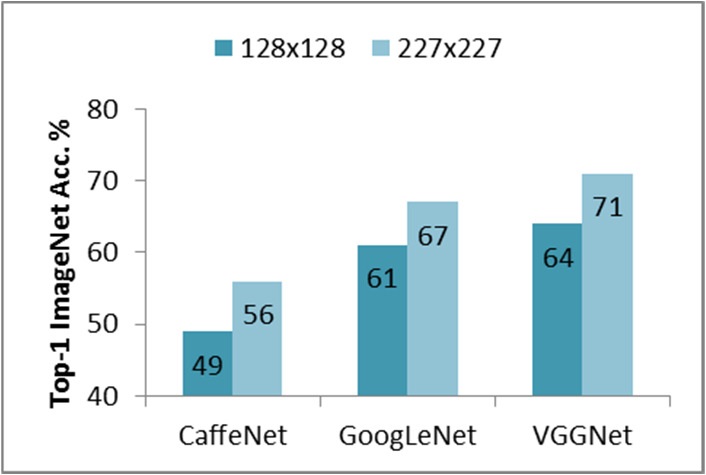}
\caption{ImageNet validation accuracy at different image resolutions for different deep networks \cite{cnnadvances}.}
\label{fig_imagenet}
\end{figure}

Current deep networks are mostly FCNs and memory requirements of such networks not only depend on the network parameters but also on input image size. For example, Inception-BN model proposed in \cite{bn} has approximately 10 million parameters with equivalent model size of 40 MB, excluding classification layer. Figure \ref{fig_gglmodel_time} illustrates the total memory and CPU time requirements needed for forward pass for different input resolution images when evaluated as a single thread for single image batch size. All timing reported in this paper are tested with Caffe framework \cite{caffe} on Intel Xeon(R) CPU E5-2640v4@2.40GHz with 128 GB RAM machine. 

As observed from Figure \ref{fig_gglmodel_time}, different input image size does reduce total memory requirement as well as computation time for same network without any architectural or parameter modifications. However, reduction in memory and computational requirements due to reduced image size normally leads to reduction in performance too as shown in Figure \ref{fig_imagenet} for ImageNet classification task. The figure illustrates the trade-off between accuracy and image resolution for different deep network architectures \cite{cnnadvances}. 
The reduction in input image size from $227 \times 227$ to $128 \times 128$ leads to a consistent drop of around 6-7\% in top-1 validation accuracy for ImageNet dataset irrespective of network architecture. 

In this paper, we focus on accuracy improvement of deep networks at low resolution images to make them deployable for face identification applications.
We utilize student-teacher paradigms proposed in \cite{net2net} and \cite{distillation} to improve performance of deep models for low resolutions images.
The focus of KD \cite{distillation} techniques is to distil knowledge in the form of ``layer representation'' from powerful teacher network to train less powerful student network, so that it outperforms itself had it been trained from scratch without knowledge distillation. In KT \cite{net2net} paradigm, the focus is to transfer knowledge in the form of ``layer parameters'' to accelerate the training of student network. By combining both these paradigms, we achieve accelerated training and inference of student network with improved performance.

\subsection{Proposed Approach}
\label{section_approach}
In our proposed methodology of student-teacher paradigm, the deep model architecture of both teacher and student network remains the same as illustrated in Figure \ref{fig_kd}. The only difference comes from input image resolution these two 
networks use while training and testing. We use smaller resolution input images for student network while teacher network uses bigger resolution images. Hence, model size of both teacher and student network remains same. However, we achieve reduction in data memory size and computational requirements due to smaller resolution images as illustrated in Figure \ref{fig_gglmodel_time} with inception-BN model architecture. 

 We propose following three strategies for training student network: first is based on using KD paradigm only, second is based on using KT paradigm only while third one is based on combining KD and KT paradigms.
 \begin{itemize}
 \item KD paradigm: extract feature representation from last global pooling layer of teacher model and use it as ''feature representation'' to train student network. Hence, we use two loss functions while training student networks: one is normal classification loss while second one is feature matching loss.
 \item KT paradigm: initialize student models with parameters of teacher model and train the initialized model with classification loss only. 
 \item Combine KD and KT: combine above two paradigms to train student network.
\end{itemize}

We train teacher and student networks on labelled face images with different resolutions $x_t \in \chi_t$ and $x_s \in \chi_s$ respectively. 
Let $\phi(x_t,\theta_t)$ and $\psi(x_s,\theta_s)$ be the feature generation operation for teacher and student networks parametrized by $\theta_t$ and $\theta_s$ respectively. We train teacher network following classification loss $L_t = L_{CS}^t$ (Eqn. \ref{eqn_tloss}) whereas student network is trained with combination of classification and feature matching loss $L_s = L_{CS}^s + \alpha L_{feat}$. (Eqn. \ref{eqn_sloss} and \ref{eqn_floss}). 

\begin{equation}
L_{CS}^{t} = - \frac{1}{|\chi_t|} \sum_{x_t \in \chi_t} log \frac{e^{W_t' \phi(x_t, \theta_t)}}{\sum_{x_t \in \chi_t}e^{W_t'\phi(x_t, \theta_t)}}
\label{eqn_tloss}
\end{equation}

\begin{equation}
L_{CS}^s =  - \frac{1}{|\chi_s|} \sum_{x_s \in \chi_s} log \frac{e^{W_s' \psi(x_s, \theta_s)}}{\sum_{x_s \in \chi_s}e^{W_s'\psi(x_s, \theta_s)}}
\label{eqn_sloss}
\end{equation}
where, $W_t$ and $W_s$ are weight matrices of the last classification layer.
We use Euclidean loss for matching features between student and teacher network:
\begin{equation}
L_{feat} = \frac{1}{|\chi_t|} \sum \Vert \phi(x_t, \theta_t) - \psi(x_s, \theta_s) \Vert_{2}^{2}
\label{eqn_floss}
\end{equation}
In $L_{feat}$ equation above, $x_t$ and $x_s$ corresponds to same image with different resolutions.
For training student networks, we use following different parameter settings:
\begin{enumerate}
\item Train from scratch: $\alpha=0$
\item Train with KD: $\alpha=0.1$, $\theta_s = random$ and $W_s = random$
\item Train with KT: $\alpha=0$, $\theta_s = \theta_t$ and $W_s = W_t$
\item Train with KT and KD: $\alpha=0.1$, $\theta_s = \theta_t$ and $W_s = W_t$
\end{enumerate}

\section{Experimental Results}
\label{section_results}

In this section, we present results for proposed deep face recognition model compression approach for shallow and deep networks.
For shallow network, we use inception-BN \cite{bn} architecture while for deep network we use $100$-layer deep residual architecture \cite{arcface}.
We evaluate performance of these networks on  LFW \cite{lfwdata} and IJB-C \cite{ijbc} datasets.
For training all networks, we used publicly available MS-Celeb1M \cite{msceleb} dataset. 
The dataset is cleaned first to remove noise and overlap between LFW and IJB-C identities.
To pre-process training and test datasets, we follow  MTCNN \cite{mtcnn} algorithm for face detection and alignment.
All evaluation results are presented for single model and single crop without any test time augmentation, unless stated otherwise. 

\subsection{Shallow Architecture}
We used inception-BN model as shallow architecture for evaluating our proposed approach. We set $224\times224$ image resolution for teacher network  whereas we used $160\times160$ and $128\times128$ as two different resolutions to train two student inception-BN models. All these models are trained with softmax loss and produce $1024$-dimension feature vector following global pooling layer. We evaluated performance of these models on LFW dataset with open-set protocol presented in \cite{lfwopenset} for single gallery image.
We used four different setting proposed in Section \ref{section_approach} to train student networks.
Table \ref{tb_ggl_acc} presents LFW open-set accuracies for these four training settings.

All techniques, KD, KT and their combination, work surprisingly well to improve accuracies of inception-BN model at low resolutions.
Biggest gain in accuracy is observed for $128\times128$ resolution in which KD, KT and their combination improves performance by 6\%, 9\% and 11\% respectively. 
We observed substantial improvements in accuracies at low resolutions with KT technique alone while improvement from KD technique is marginal.
Hence, KT technique not only accelerates training but also helps for accuracy improvements for face recognition applications. 

\begin{table}[t!]
\centering
\begin{tabular}{ |c|c|c|c|c| } 
 \hline
 Resolution & Train from & KD & KT & KD+KT \\ 
  & scratch &  &  &  \\ 
 \hline
 \hline
 224$\times$224 & 89.09 & - & - & - \\
 \hline
 160$\times$160  & 85.91 & 87.75 & 88.42 & 88.76 \\
 \hline
 128$\times$128 & 76.68 & 82.21 & 85.74 & 86.24 \\
 \hline
\end{tabular}\\
\caption{LFW open-set accuracies (DIR(\%)@FAR=1\%) of inception-BN model for different resolutions and knowledge transfer techniques. Model trained with 224$\times$224 model.}
\label{tb_ggl_acc}
\end{table}

\begin{figure}[t!]
\centering
\includegraphics[height=1.5in]{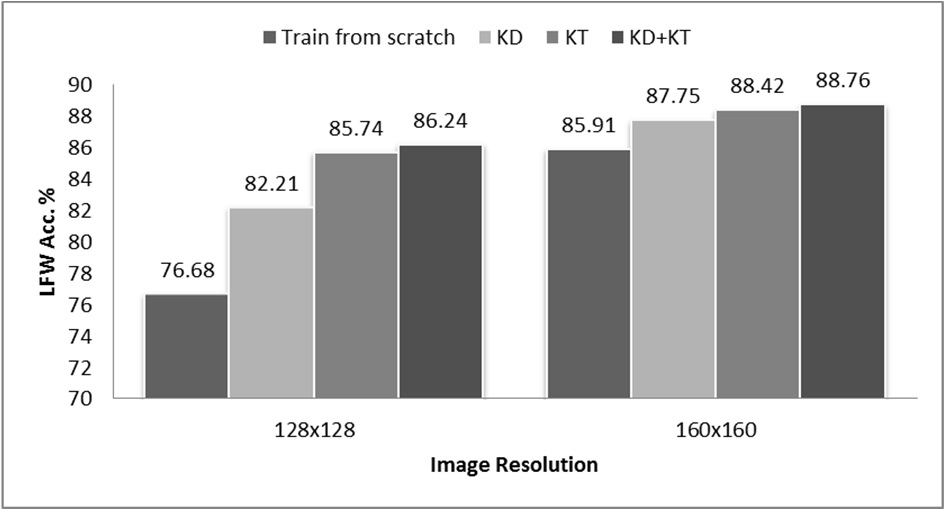}
\caption{Accuracy improvements on LFW open-set accuracies (DIR(\%)@FAR=1\%) of inception-BN model for different resolutions and knowledge transfer techniques.}
\label{fig_ggl_acc}
\end{figure}

\subsection{Deep Architecture}
\subsubsection{Training Details}
For evaluation of deep architecture, we used 100-layer residual model proposed in \cite{arcface}.
We further incorporated  modifications to above architecture pertaining to face recognition domain.
We set $112\times112$ image resolution for teacher network whereas different resolutions of $96\times96$, $80\times80$, $64\times64$ and $48\times48$ are used to train different student networks. 
All networks are trained with ArcFace loss \cite{arcface} with $256$-dimension feature embedding as final layer feature representation.

\subsubsection{Testing Dataset}
The IJB-C \cite{ijbc} is the most recent dataset collected under unconstrained settings of extreme viewpoints, resolution and illumination variations making it the most challenging than the commonly used LFW dataset.
The dataset contains $3531$ identities and $163,359$ images and video frames. 
The dataset has defined three protocols for evaluating verification and identification performances. 
IJB-C 1:1 mixed verification protocol contains $19,557$ genuine and $15,638,932$ imposter pairs for comparisons.
IJB-C 1:1 covariate verification protocol has $47,404,001$ pair of templates ($7,819,362$ genuine and $39,584,639$ imposter) created from $140,739$ images and video frames.
Moreover, IJB-C 1:N mixed identification protocol contains two splits and we report average results of these two splits.

\begin{table}[t!]
\centering
\begin{tabular}{|c|c|c|} 
 \hline
 & \multicolumn{2}{|c|}{IJB-C 1:1 TAR (in \%)@FAR} \\
 \hline
 Method & \hspace{5mm} $10^{-4}$ \hspace{5mm} & $10^{-5}$ \\ 
 \hline
 \hline
 Crystal Loss \cite{crystalloss} & 92.50 & 87.75 \\ 
 \hline
 P2SGrad \cite{P2SGrad} & 92.25 & 87.84 \\ 
 \hline
 Fixed AdaCos \cite{adacos} & 92.35 & 87.87 \\ 
 \hline
 Dynamic AdaCos \cite{adacos} & 92.40 & 88.03 \\ 
 \hline
 ArcFace \cite{arcface} & 95.65 & 93.15 \\ 
 \hline
 \bf{Ours} & \bf{96.39} & \bf{94.20} \\
 \hline
\end{tabular} \\
\caption{Evaluation results on IJB-C 1:1 mixed verification protocol of our 100-layer teacher network trained for $112\times112$ image resolution with cleaned and overlap removed MS-Celeb1M dataset \cite{msceleb}. We use MTCNN for face detection and alignment.}
\label{tb_ijbc11_mix}
\end{table}

\subsubsection{Teacher Results}
Evaluation results of teacher network on three protocols of IJB-C dataset are presented in Tables \ref{tb_ijbc11_mix}, \ref{tb_ijbc1n_mix}  and \ref{tb_ijbc11_covariate} with MTCNN. 
Table \ref{tb_ijbc11_mix} and \ref{tb_ijbc1n_mix} presents results on IJB-C 1:1 mixed verification and identification protocols.
Mixed protocol contains multiple medias for each template. Final template representation was obtained by simple average pooling of individual media feature. 
Table \ref{tb_ijbc11_covariate} presents results on IJB-C 1:1 covariate verification protocol. This protocol uses single media image for each template and specifically designed to evaluate performance of model under different  covariate conditions.
Our teacher network achieves top-1 performance on all three protocols of IJB-C dataset with single deep model and single crop.

\begin{table*}[t!]
\centering
\begin{tabular}{|c|c|c|c|c|c|} 
 \hline
 & \multicolumn{5}{|c|}{IJB-C 1:N Identification} \\
  \hline
 & \multicolumn{3}{|c|}{TPIR (in \%)@FPIR} & \multicolumn{2}{|c|}{Retrieval Rate (in \%)} \\
 \hline
 Method & \hspace{1mm} 0.001 \hspace{1mm} & \hspace{1mm} 0.01 \hspace{1mm} & \hspace{1mm} 0.1 \hspace{1mm} & \hspace{1mm} Rank=1 \hspace{1mm} & \hspace{1mm} Rank=10 \hspace{1mm}\\ 
  \hline
  \hline
 Crystal Loss \cite{crystalloss} & 78.54 & 87.01 & 92.10 & 94.57 & 97.48  \\
\hline
 \bf{Ours} & \bf{95.51} & \bf{96.05} & \bf{97.07} & \bf{97.99} & \bf{98.86} \\
 \hline
\end{tabular} \\
\caption{Evaluation results of teacher network on IJB-C 1:N mixed identification protocol with MTCNN and average pooling.}
\label{tb_ijbc1n_mix}
\end{table*}

\begin{table*}[t!]
\centering
\begin{tabular}{ |c|c|c|c|c|c|c|c| } 
 \hline
 & \multicolumn{7}{|c|}{IJB-C 1:1 Covaraite TAR (in \%)@FAR} \\
 \hline
 Method & \hspace{1mm} $10^{-1}$ \hspace{1mm} & \hspace{1mm} $10^{-2}$ \hspace{1mm} & \hspace{1mm} $10^{-3}$ \hspace{1mm} & \hspace{1mm} $10^{-4}$ \hspace{1mm} & \hspace{1mm} $10^{-5}$ \hspace{1mm} & \hspace{1mm} $10^{-6}$ \hspace{1mm} & \hspace{1mm} $10^{-7}$ \hspace{1mm}\\ 
 \hline
 \hline
 \cite{covariate} & 93.68 & 86.24 & 76.60 & 64.03 & 50.23 & 35.96 & 24.17 \\ 
 \hline
 Fusion \cite{covariate} & 96.81 & 92.61 & 85.99 & 76.23 & 64.78 & 52.49 & 23.71 \\
 \hline
\bf{Ours} & \bf{96.09}	& \bf{93.81}	& \bf{90.93}	& \bf{84.08} &	\bf{69.93}	& \bf{53.66} & \bf{31.62} \\
 \hline
\end{tabular}\\
\caption{Evaluation results of teacher network on IJB-C 1:1 covariate protocol with MTCNN.}
\label{tb_ijbc11_covariate}
\end{table*}

\begin{table*}[t!]
\centering
\begin{tabular}{|c|c|c|c|c|c|} 
 \hline
 & \multicolumn{5}{|c|}{IJB-C 1:N Identification} \\
  \hline
 & \multicolumn{3}{|c|}{TPIR (in \%)@FPIR} & \multicolumn{2}{|c|}{Retrieval Rate (in \%)} \\
 \hline
 Method & \hspace{1mm} 0.001 \hspace{1mm} & \hspace{1mm} 0.01 \hspace{1mm} & \hspace{1mm} 0.1 \hspace{1mm} & \hspace{1mm} Rank=1 \hspace{1mm} & \hspace{1mm} Rank=10 \hspace{1mm}\\ 
  \hline
  \hline
 MTCNN & 95.51 & 96.05 & 97.07 & 97.99 & 98.86 \\
 \hline
 MTCNN+Flip & 95.62 & 96.12 & 97.23 & 98.02 & 98.84 \\
 \hline
 MTCNN+Flip+Score & 95.76 & 96.26 & 97.31 & 98.11 & 98.83 \\
 \hline
 RetinaFace & 95.85 & 96.29 & 97.34 & 98.22 & 98.93 \\
 \hline
 RetinaFace+Flip & 96.04 & 96.59 & 97.50 & 98.38 & 98.97 \\
 \hline
 RetinaFace+Flip+Score & 96.11 & 96.62 & 97.50 & 98.39 & 98.95 \\
 \hline
\end{tabular} \\
\caption{Evaluation results of teacher network on IJB-C 1:N mixed identification protocol for different face detector and alignment methods with average and FD score based feature fusion.}
\label{tb_ijbc1n_mix_compare} 
\end{table*}

\begin{table}[h!]
\centering
\begin{tabular}{|c|c|c|c|} 
 \hline
 & \multicolumn{3}{|c|}{IJB-C 1:1 TAR (in \%)@FAR} \\
 \hline
 Method & \hspace{1mm} $10^{-4}$ \hspace{1mm} & \hspace{1mm} $10^{-5}$ \hspace{1mm} & $10^{-6}$ \\ 
 \hline
 \hline
 MTCNN & 96.39 & 94.20 & 89.73\\
 \hline
 MTCNN+Flip & 96.38 & 94.44 & 89.06\\
 \hline
 MTCNN+Flip+Score & 96.66 & 94.59 & 89.55\\
 \hline
 RetinaFace & 96.65 & 94.48 & 89.23\\
 \hline
 RetinaFace+Flip & 96.81 & 94.79 & 88.81\\
 \hline
 RetinaFace+Flip+Score & 96.86 & 94.95 & 88.92\\
 \hline
\end{tabular} \\
\caption{Evaluation results on IJB-C 1:1 mixed verification protocol for different face detector and alignment methods with average and FD score based feature fusion.}
\label{tb_ijbc11_mix_compare}
\end{table}

Performance of teacher network with RetinaFace \cite{retinaface}, the recently proposed unified face detector and alignment approach, are presented in Tables \ref{tb_ijbc1n_mix_compare} and \ref{tb_ijbc11_mix_compare} for IJB-C mixed protocols.
We observed improvements in accuracies by replacing MTCNN with RetinaFace. Further accuracy improvements are obtained by doing test time augmentation such as image flipping and feature averaging. Furthermore, constructing templates by replacing average feature pooling with weighted feature pooling with weights obtained from face detector stage improves accuracy of IJB-C mixed protocols. 

\begin{table}[h!]
\centering
\begin{tabular}{ |c|c|c|c|c| } 
 \hline
  & \multicolumn{4}{|c|}{IJB-C 1:1 TAR (in \%)@FAR} \\
  \hline
  & \multicolumn{2}{|c|}{w/o KT} & \multicolumn{2}{|c|}{with KT} \\
  \hline
 Resolution & \hspace{1mm} $10^{-4}$ \hspace{1mm} & \hspace{1mm} $10^{-5}$ \hspace{1mm} & \hspace{1mm} $10^{-4}$ \hspace{1mm} & \hspace{1mm} $10^{-5}$ \hspace{1mm} \\ 
 \hline
 \hline
 \bf{112$\times$112} & \bf{96.39} & \bf{94.20} & - & - \\
 \hline
 \bf{96$\times$96} & \bf{95.89} & \bf{93.41} & \bf{96.05} & \bf{93.66} \\
 \hline
 \bf{80$\times$80}  & 95.45& 92.46 & \bf{96.01} & \bf{93.19} \\
 \hline
 64$\times$64 & 94.81 & 91.51 & 95.67 & 92.55 \\
 \hline
 48$\times$48 & 91.74 & 86.17 & 93.94 & 89.57 \\
 \hline
\end{tabular}\\
\caption{Knowledge transfer results for IJB-C 1:1 mixed verification protocol at different resolutions with MTCNN and average pooling.}
\label{tb_ijbc11_mix_kt}
\end{table}

\begin{figure*}[t!]
    \centering
    \begin{subfigure}[t]{0.4\textwidth}
        \centering
        \includegraphics[width=\textwidth]{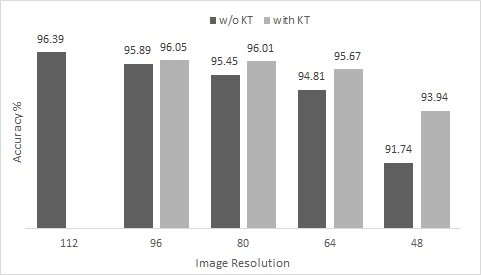}
        \caption{$10^{-4}$}
    \end{subfigure} 
  \hspace{10mm}
    \begin{subfigure}[t]{0.4\textwidth}
        \centering
        \includegraphics[width=\textwidth]{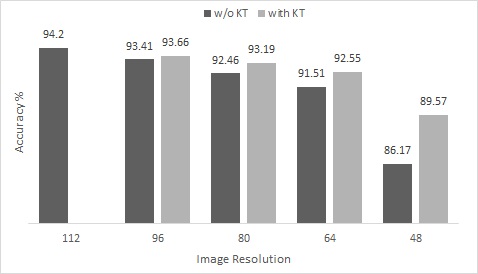}
       \caption{$10^{-5}$}
    \end{subfigure}
    \caption{Accuracy results for IJB-C 1:1 mixed verification protocol at (a) $10^{-4}$ and (b) $10^{-5}$ FAR for student networks trained from scratch and with KT.}
    \label{fig_ijbc11_mix_kt}
\end{figure*}

\begin{table*}[t!]
\centering
\begin{tabular}{|c|c|c|c|c|c|c|c|c|c|c|} 
 \hline
 & \multicolumn{10}{|c|}{IJB-C 1:N Identification} \\
 \hline 
  & \multicolumn{5}{|c|}{w/o KT} & \multicolumn{5}{|c|}{with KT} \\
  \hline
 & \multicolumn{3}{|c|}{TPIR (in \%)@FPIR} & \multicolumn{2}{|c|}{Retrieval Rate (in \%)} & \multicolumn{3}{|c|}{TPIR (in \%)@FPIR} & \multicolumn{2}{|c|}{Retrieval Rate (in \%)}\\
 \hline
 Resolution & 0.001 & 0.01 & 0.1 & Rank=1 & Rank=10 & 0.001 & 0.01 & 0.1 & Rank=1 & Rank=10\\ 
 \hline
  \hline
112$\times$112 & 95.51 & 96.05 & 97.07 & 97.99 & 98.86 & - & - & - & - & - \\
\hline
 96$\times$96 & 94.66 & 95.32 & 96.66 & 97.74 & 98.71 & 94.87 & 95.51 & 96.81 & 97.96 & 98.77 \\
 \hline
 80$\times$80 & 93.16 & 94.11 & 95.99 & 97.24 & 98.53 & 94.48 & 95.26 & 96.79 & 97.75 & 98.7 \\
 \hline
 64$\times$64 & 92.41 & 93.41 & 95.65 & 97.15 & 98.48 & 93.7 & 94.61 & 96.53 & 97.55 & 98.66  \\
 \hline
 48$\times$48 & 86.25 & 88.32 & 92.41 & 95.16 & 97.67 & 90.97 & 92.19 & 94.66 & 96.59 & 98.29 \\
 \hline
\end{tabular} \\
\caption{Knowledge transfer results for IJB-C 1:N Identification protocol at different resolutions with MTCNN and average pooling.}
\label{tb_ijbc1n_mix_kt}
\end{table*}

\subsubsection{Student Results}
We trained several students networks at different resolutions from scratch and with KT technique.
We did not use KD technique while training deep student networks as performance improvement from KD was marginal for shallow architectures.
The evaluation results of these student networks on IJB-C mixed protocols are presented in Tables \ref{tb_ijbc11_mix_kt} and \ref{tb_ijbc1n_mix_kt}.
We observed substantial improvement in accuracies for both protocols of IJB-C dataset for student networks following KT technique.
As illustrated in Figure \ref{fig_ijbc11_mix_kt}, more accuracy improvements are observed at lower resolutions as compared with bigger resolutions.
Student networks trained following KT technique at $96\times96$ and $80\times80$ resolutions even achieve top-1 accuracy results  on IJB-C 1:1 mixed verification protocol. Student networks at $64\times64$ and $48\times48$ resolutions also show significant improvement in IJB-C mixed protocol accuracies. Similarly, all student networks also achieve state-of-the-art accuracies on IJB-C 1:N mixed identification protocol as presented in Table \ref{tb_ijbc1n_mix_kt}. All results are obtained following with MTCNN and simple average feature pooling method.

Table \ref{tb_student_time} reports model size, inference time and total number of multiply-accumulate (MACC) operations for all deep networks.
The students network trained at different resolutions offer different trade-off in terms of accuracy, inference time and memory usage.
Student network trained at $80\times80$ resolution reduces CPU time requirement and number of MACC operations by half as compared to teacher network while maintaining almost same accuracies on IJB-C dataset.
Following Table \ref{tb_student_time}, one can choose appropriate student architecture depending on deployment constraints.

\begin{table}[t!]
\centering
\begin{tabular}{ |c|c|c|c| } 
 \hline
 Resolution & Model size & CPU time & \# MACC\\ 
 & in MB & in ms & in giga \\  
 \hline
 \hline
 112$\times$112 & 224 & 517 & 12.1  \\
 \hline
 96$\times$96 & 217 & 381 & 8.89 \\
 \hline
 80$\times$80  & 212 & 251 & 6.17 \\
 \hline
 64$\times$64 & 207 & 182 & 3.95 \\
 \hline
 48$\times$48 & 204 & 115 & 2.22 \\
 \hline
\end{tabular}\\
\caption{Model size, inference time and number of MACC operations for 100-layer deep model for different input image resolutions.}
\label{tb_student_time}
\end{table}

\section{Conclusion}
In this paper, we proposed alternative approach for deep face model compression by combining knowledge transfer and distillation paradigms.
Performing simultaneous knowledge distillation and transfer from high resolution teacher to low resolution student network provides fourfold advantage of accelerated training, accelerated testing, reduced memory requirements and huge accuracy improvements.
The proposed approach provides nice trade-off between computational complexity and accuracy to choose appropriate model based on final 
hardware constraints.  The reduction in computational complexity is achieved by neither modifying the network architecture nor pruning/quantizing model parameters, making it easier to train and deploy student network than other deep model compression techniques introduced in literature.
However, further compressing student model by combining these existing techniques will be explored in future.

{\small
\bibliographystyle{ieee_fullname}
\bibliography{facebib2019}
}

\end{document}